\documentclass[10pt,twocolumn,letterpaper]{article}

\usepackage{cvpr}              

\usepackage[table]{xcolor}
\usepackage{graphicx}
\usepackage{amsmath}
\usepackage{amssymb}
\usepackage{booktabs}
\usepackage{multirow}
\usepackage{makecell}
\usepackage{bm}
\usepackage{enumerate}
\usepackage{arydshln}
\usepackage{booktabs}
\usepackage{multirow}
\usepackage{tabularx}
\usepackage{algorithmic}
\usepackage{algorithm}
\usepackage{amsthm}
\usepackage{color}
\usepackage{xfrac}
\usepackage{threeparttable}
\usepackage{siunitx}
\usepackage{caption}
\usepackage{hhline}

\usepackage[pagebackref,breaklinks,colorlinks]{hyperref}
\usepackage[symbol]{footmisc}
\definecolor{best}{RGB}{255, 153, 153}
\definecolor{second}{RGB}{255, 204, 153}
\newcolumntype{C}{>{\centering\arraybackslash}X}

\usepackage[capitalize]{cleveref}
\crefname{section}{Sec.}{Secs.}
\Crefname{section}{Section}{Sections}
\Crefname{table}{Table}{Tables}
\crefname{table}{Tab.}{Tabs.}

\def\cvprPaperID{6101} 
\def\confName{CVPR}
\def\confYear{2022}

\begin{document}

\title{Hallucinated Neural Radiance Fields in the Wild}

\author{Xingyu Chen$^{1}$\footnote{} \qquad Qi Zhang$^{2}$\footnote{} \qquad Xiaoyu Li$^{2}$ \qquad Yue Chen$^{1}$ \\
\qquad Ying Feng$^2$ \qquad  Xuan Wang$^2$ \qquad Jue Wang$^2$ \vspace{3pt}\\
$^{1}$Xi'an Jiaotong University\ \ \ \  $^{2}$Tencent AI Lab\\
}

\twocolumn[{%
\renewcommand\twocolumn[1][]{#1}%
\maketitle
\vspace{-0.5cm}
\begin{center}
    \includegraphics[width=1\hsize]{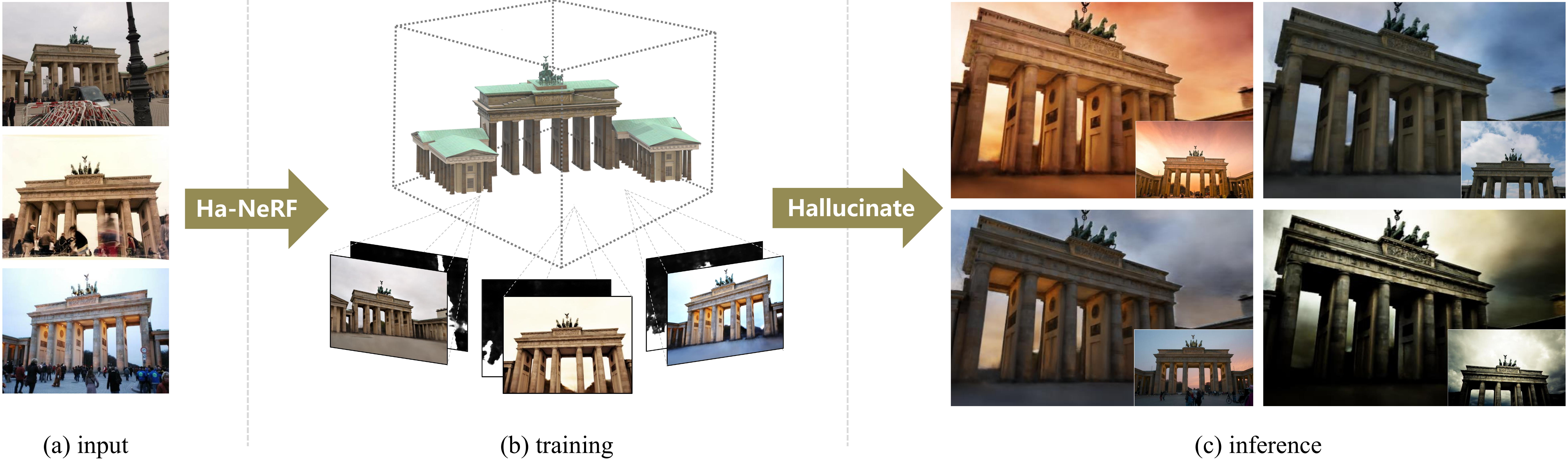}
    \captionof{figure}{We recover (b) hallucinated neural radiance fields (Ha-NeRF) from (a) a group of tourism images with variable appearance and complex occlusions. Our method can consistently render (c) free-occlusion views which hallucinate different appearances. }
    \label{fig:motivation}
\end{center}
}]

\footnotetext[1]{Work done during an internship at Tencent AI Lab.}
\footnotetext[2]{Corresponding Author.}

\begin{abstract}

    Neural Radiance Fields (NeRF) has recently gained popularity for its impressive novel view synthesis ability. This paper studies the problem of hallucinated NeRF: i.e., recovering a realistic NeRF at a different time of day from a group of tourism images. Existing solutions adopt NeRF with a controllable appearance embedding to render novel views under various conditions, but they cannot render view-consistent images with an unseen appearance. To solve this problem, we present an end-to-end framework for constructing a hallucinated NeRF, dubbed as Ha-NeRF. Specifically, we propose an appearance hallucination module to handle time-varying appearances and transfer them to novel views. Considering the complex occlusions of tourism images, we introduce an anti-occlusion module to decompose the static subjects for visibility accurately. Experimental results on synthetic data and real tourism photo collections demonstrate that our method can hallucinate the desired appearances and render occlusion-free images from different views.  The project and supplementary materials are available at \href{https://rover-xingyu.github.io/Ha-NeRF/}{https://rover-xingyu.github.io/Ha-NeRF/}.
 
\end{abstract}
\section{Introduction}
\label{sec:intro}

In recent years, synthesizing photo-realistic novel views of a scene has become a research hotspot along with the rapid development of neural rendering technologies. Imagine you want to visit the Brandenburg Gate in Berlin and enjoy the landscapes at different times and weathers, but you cannot because of the coronavirus pandemic. For this hallucinated experience to be as engaging as possible, photo-realistic images from different views that can change with the weather, time, and other factors are necessary.

To achieve this, Neural Radiance Fields (NeRF) \cite{mildenhall2020nerf} and its following methods \cite{yu2021pixelnerf,li2021neural,pumarola2021d} have shown a remarkable capacity to recover the 3D geometry and appearance, giving the user an immersive feeling of physically being there. However, one significant drawback of NeRF is that they require a group of images without variable illumination and moving objects, \ie, the radiance of the scene is constant and visible for each view. Unfortunately, most images of tourist landmarks are internet photos captured at different times and occluded by various objects. Most NeRF-based methods would integrate variable appearances and transient occluders into the 3D volume when they occur, which disturbs the real scene in the volume. How to synthesize the occlusion-free views from images with variable appearances and occluders remains to be solved.

Martin-Brualla \etal~\cite{martin2021nerf} attempt to tackle the aforementioned problem by proposing a NeRF in the Wild method (NeRF-W). They optimize an appearance embedding for each input image to address variable appearances and use a transient volume to decompose static components and their occlusion. Compared to NeRF, NeRF-W takes a step towards recovering a realistic world from tourism images with variable appearances and occluders.
However, NeRF-W implements a controllable appearance by the optimized embeddings from train samples, making it need to optimize the embeddings when given a new image and can not hallucinate an appearance from other datasets.
Furthermore, NeRF-W tries to optimize a transient volume for each input image with a transient embedding as input, which is highly ill-posed due to the randomness of transient occluders. And this leads to the inaccurate decomposition of the scene and further causes the entanglement of appearances and occlusion, \eg, results in the transient volume to remember the sunset glow.

To address these limitations, we present a hallucinated NeRF (Ha-NeRF) framework that can hallucinate the realistic radiance field from unconstrained tourist images with variable appearances and occluders, as shown in Fig. \ref{fig:motivation}. For appearance hallucination, we propose a CNN-based appearance encoder and a view-consistent appearance loss to transfer consistent photometric appearance in different views. This design gives our method the flexibility to transfer the appearance of unlearned images. For anti-occlusion, we utilize an MLP to learn an image-dependent 2D visibility mask with an anti-occlusion loss that can automatically separate the static components with high accuracy during training. Experiments on several landmarks confirm the superior of the proposed method in terms of appearance hallucination and anti-occlusion.

Our contributions can be summarized as follows:
\begin{enumerate}
    \item The Ha-NeRF is proposed to recover the appearance hallucination radiance fields from a group of images with variable appearances and occluders. 
    \item An appearance hallucination module is developed to transfer the view-consistent appearance to novel views.
    \item An anti-occlusion module is modeled image-dependently to perceive the ray visibility. 
\end{enumerate}

\section{Related Work}
\noindent\textbf{Novel View Synthesis.}
Rendering photo-realistic images is at the heart of computer vision and has been the focus of decades of research. Traditionally, view synthesis could be considered as an image-based warping task combined with geometry structure~\cite{shum2000review}, such as implicit geometry from dense images~\cite{mcmillan1995plenoptic, levoy1996light, gortler1996lumigraph, buehler2001unstructured, davis2012unstructured} and explicit geometry~\cite{debevec1996modeling, hedman2017casual, hedman2018instant, cayon2015bayesian, overbeck2018system}. Recent works have used a set of unconstrained photo collections to explicitly infer the light and reflectance of the objects in the scene \cite{laffont2012coherent, shan2013visual}. Others make use of semantic information to restore transient objects \cite{price2018augmenting}.

With the advancement of deep learning, many approaches have applied deep learning techniques to improve the performance of view synthesis. Researchers try to combine convolutional neural networks with scene geometry to predict depth or planar homography for novel view synthesis~\cite{zhou2016view, liu2018geometry, niklaus20193d, choi2019extreme, wiles2020synsin, hedman2018deep}. Inspired by the layered depth images \cite{shade1998layered}, recent works exploit explicit scene representation (\eg, multi-plane images, multiple sphere images) and render novel views using alpha-compositing for novel view synthesis \cite{zhou2018stereo, srinivasan2019pushing, mildenhall2019local,flynn2019deepview, broxton2020immersive, tucker2020single}.
More recently, researchers have focused on the challenging problem of learning implicit functions (\eg, encoded features, NeRF) to represent scenes for novel view synthesis \cite{mildenhall2020nerf,yu2021pixelnerf, riegler2020free,riegler2021stable}.

\noindent\textbf{Neural Rendering.}
Neural rendering \cite{tewari2020state} is closely related and combines ideas from classical computer graphics and deep learning to create algorithms for synthesizing image and reconstruction geometry from real-world observations. Several works present different ways to inject learning components into the rendering pipeline, such as learned latent textures \cite{thies2019deferred}, point clouds \cite{dai2020neural,aliev2020neural}, occupancy fields \cite{michalkiewicz2019implicit}, signed distance functions \cite{park2019deepsdf}. 
Based on the image translation network, Meshry \etal \cite{meshry2019neural} learned a neural re-rendering network conditioned on a learned latent appearance embedding module to recover point cloud for view synthesis. However, the utilization of an image translation network leads to temporal artifacts visible under camera motion.

With the development of volume rendering \cite{lombardi2019neural, mildenhall2020nerf,sitzmann2019scene}, it is easy to render realistic and consistent views. Mildenhall \etal \cite{mildenhall2020nerf} propose Neural Radiance Fields (NeRF) and use a multi-layer perceptron (MLP) to restore a radiance field. Many following works try to extend NeRF to the dynamic scene \cite{li2021neural,pumarola2021d,chen2021mvsnerf,yu2021pixelnerf}, fast training and rendering \cite{cole2021differentiable,garbin2021fastnerf,reiser2021kilonerf,yu2021plenoctrees} and scene edit \cite{martin2021nerf,barron2021mip,niemeyer2021giraffe,zhang2021editable}. Martin-Brualla \etal \cite{martin2021nerf} propose NeRF in the wild (NeRF-W) to optimize the appearance and tackle occlusion via static volume and dynamic volume respectively, but they failed in some scenes. Their dynamic volume is often used to describe the dramatic changes in appearance, such as view-dependent lighting. Besides, while NeRF-W implements a controllable appearance, it is hard to hallucinate consistent views at an appearance that has never been seen.

\noindent\textbf{Appearance Transfer.}

A given scene can take on dramatically diverse appearances in different weather conditions and at different times. Grag \etal \cite{garg2009dimensionality} propose that the dimensionality of scene appearance in tourist images captured at the same position is relatively low, except for outliers like transient objects. One can recover appearance for a photo collection by estimating coherent albedos across the collection \cite{laffont2012coherent}, isolating surface albedo and scene illumination from the shape recovery \cite{kim2016multi}, retrieving the sun's location through timestamps and geolocation \cite{hauagge2014reasoning}, or assuming a fixed view \cite{sunkavalli2007factored}. However, these methods assume simple lighting models that do not apply to nighttime scene appearance. Radenovic \etal \cite{radenovic2016dusk} restore distinct day and night reconstructions, but are unable to achieve a smooth gradation of appearance from day to night. Park \etal \cite{park2016efficient} propose an efficient technique to optimize the appearance of a collection of images depicting a common scene. Meshry \cite{meshry2019neural} uses a data-driven implicit representation of appearance that is learned from the input image distribution, while Martin-Brualla \etal \cite{martin2021nerf} extend the data-driven method to NeRF and optimize appearance latent code for each view for appearance controllable. In contrast, the proposed method tries to learn the appearance features that are decomposed from views, which means it could consistently hallucinate novel views at an unlearnt appearance.


\section{Preliminary}

We first introduce Neural Radiance Fields (NeRF) \cite{mildenhall2020nerf} that Ha-NeRF extends. NeRF represents a scene using a continuous volumetric function $F_{\theta}$ that is modeled as a multilayer perceptron (MLP). It takes a 3D location $\mathbf{x}=(x, y, z)$ and 2D viewing direction $\mathbf{d}=(\alpha, \beta )$ as input and output an emitted color $\mathbf{c}=(r, g, b)$ and volume density $\sigma$ as:
\begin{equation}\small
\label{eq1}
\begin{split}\small
(\sigma, \mathbf{z}) &= F_{\theta_{1}}(\gamma_{\mathbf{x}}(\mathbf{x})), \\
\mathbf{c}           &= F_{\theta_{2}}(\gamma_{\mathbf{d}}(\mathbf{d}), \mathbf{z}),
\end{split}
\end{equation}
where $\theta=(\theta_{1}, \theta_{2})$ are the MLP parameters, $\gamma_{\mathbf{x}}(\cdot)$ and $\gamma_{\mathbf{d}}(\cdot)$ are the positional encoding functions that are applied to each of the values in $\mathbf{x}$ and $\mathbf{d}$ respectively. To render the color of a ray passing through the scene, NeRF approximates the volume rendering integral using numerical quadrature. Let $\mathbf{r}(t)=\mathbf{o}+t \mathbf{d}$ be the ray emitted from the camera center $\mathbf{o}$ through a given pixel on the image plane. The approximation of the color $\hat{\mathbf{C}}(\mathbf{r})$ of the pixel is:
\begin{equation}\small
\label{eq2}
\begin{split}\small
\hat{\mathbf{C}}(\mathbf{r}) &= \sum_{k=1}^{K} T_{k} (1 - \exp(-\sigma_{k}\delta_{k}))\mathbf{c}_{k}, \\
                       T_{k} &= \exp (-\sum_{l=1}^{k-1} \sigma_{l}\delta_{l}),
\end{split}
\end{equation}
where $\mathbf{c}_k$ and $\sigma_k$ are the color and density at point $\mathbf{r}(t_k)$, $\delta_{k}=t_{k+1}-t_{k}$ is the distance between two quadrature points. Stratified sampling is used to select quadrature points $\{t_{k}\}_{k=1}^{K}$ between the near and far planes of the camera. Intuitively, alpha compositing with alpha values $1 - \exp(-\sigma_{k}\delta_{k})$ can be interpreted as the probability of a ray terminating at the location $\mathbf{r}(t_k)$, and function $T_k$ corresponds to the accumulated transmittance along the ray from the near plane to $\mathbf{r}(t_k)$.

 To optimize the MLP parameters, NeRF minimizes the sum of squared errors between an image collection $\left\{\mathcal{I}_{i}\right\}_{i=1}^{N}$ and the corresponding rendered output. Each image $\mathcal{I}_{i}$ is registered with its intrinsic and extrinsic camera parameters which can be estimated using structure-from-motion algorithms. NeRF precomputes the set of camera rays $\{\mathbf{r}_{i j}\}$ at pixel $j$ from image $\mathcal{I}_{i}$ with each ray $\mathbf{r}_{i j}(t)=\mathbf{o}_{i}+t \mathbf{d}_{i j}$ passing through the 3D location $\mathbf{o}_{i}$ with direction $\mathbf{d}_{i j}$. All parameters are optimized by minimizing the following loss:

\begin{equation}\small\label{eq5}
\mathcal{L}=\sum_{ij}\left\|\mathbf{C}(\mathbf{r}_{ij})-\hat{\mathbf{C}}(\mathbf{r}_{ij})\right\|_{2}^{2},
\end{equation}
where $\mathbf{C}\left(\mathbf{r}_{ij}\right)$ is the observed color of ray $j$ in image $\mathcal{I}_{i}$.

\section{Method}
\begin{figure}[t]
\centering

  \includegraphics[width=1\linewidth]{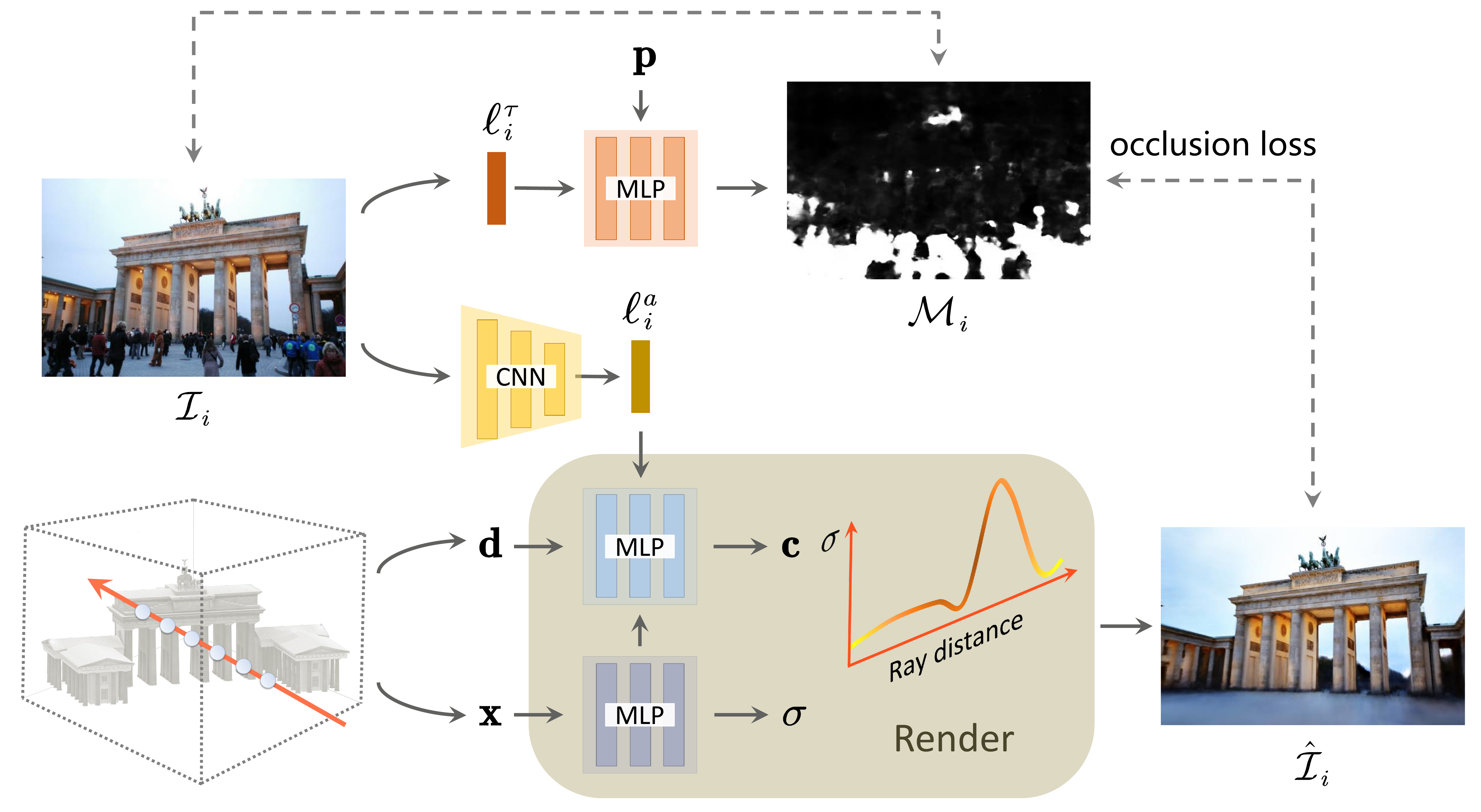}
  \vspace{-0.4cm}
   \caption{An overview of the Ha-NeRF architecture. Given an image $\mathcal{I}_{i}$, we use a CNN to encode it into an appearance latent vector $\ell_{i}^{a}$. We synthesize images by sampling location $\mathbf{x}$ and viewing direction $\mathbf{d}$ of camera rays, feeding them with $\ell_{i}^{a}$ into MLPs to produce a color $\mathbf{c}$ and volume density $\sigma$ and rendering a reconstructed image $\hat{\mathcal{I}_{i}}$. Given an image-dependent transient embedding $\ell_{i}^{\tau}$, we use an MLP to map pixel location $\mathbf{p}$ to a visible possibility $\mathcal{M}_i$, so that we can disentangle static and transient phenomena of the images with an occlusion loss.}
  \vspace{-0.4cm}
\label{fig:architecture}
\end{figure}

Given a photo collection of a scene with varying appearances and transient occluders, we aim to reconstruct the scene that can be hallucinated from a new shot while handling the occlusion. That's to say that we can modify the appearance of the whole 3D scene according to a new view captured at a different photometric condition. More specifically, taking a photo in the wild as input, we reconstruct an appearance-independent NeRF modulated by an appearance embedding encoded by a convolutional neural network in Sec.~\ref{sec:apperance}. To address the transient occluders in the photo, we propose an occlusion handling module to separate the static scene automatically in Sec.~\ref{sec:occlusion}. Fig.~\ref{fig:architecture} illustrates the overview of the proposed architecture. Next, we subsequently elaborate on each module.

\subsection{View-consistent Hallucination}
\label{sec:apperance}

To achieve the hallucination of a 3D scene according to a new shot from the input with varying appearances, the core problems are how to disentangle the scene geometry from appearances and how to transfer the new appearance to the reconstructed scene. NeRF-W~\cite{martin2021nerf} tries to use an optimized appearance embedding to explain the image-dependent appearances in the input. However, this embedding needs to be optimized during training, making it need to optimize the embeddings for hallucinating the scene from a new shot beyond the training samples and can not hallucinate an appearance from other datasets.

Therefore, we propose to learn the disentangled appearance representations using a convolutional neural network based encoder $E_{\phi}$, of which parameters $\phi$ account for the varying lighting and photometric postprocessing in the input. $E_{\phi}$ encodes each image $\mathcal{I}_{i}$ into an appearance latent vector $\ell_{i}^{a}$. The radiance $\mathbf{c}$ in Eq.~\ref{eq1} is extended to an appearance-dependent radiance $\mathbf{c}^{\ell_{i}^{a}}$, which introduces a dependency on appearance latent vector $\ell_{i}^{a}$ to emitted color:
\begin{equation}\small\label{eq7}
\mathbf{c}^{\ell_{i}^{a}}=F_{\theta_{2}}\left(\gamma_{\mathbf{d}}(\mathbf{d}), \mathbf{z},  \ell_{i}^{a}\right), \text { where } \ell_{i}^{a} = E_{\phi}(\mathcal{I}_{i}).
\end{equation}
The parameters $\phi$ of appearance encoder $E_{\phi}$ are learned alongside parameters $\theta$ of radiance field $F_{\theta}$. This appearance encoder enables our method to have the flexibility to use the appearance of images beyond the training set.

\begin{figure}[t]
\centering

  \includegraphics[width=1\linewidth]{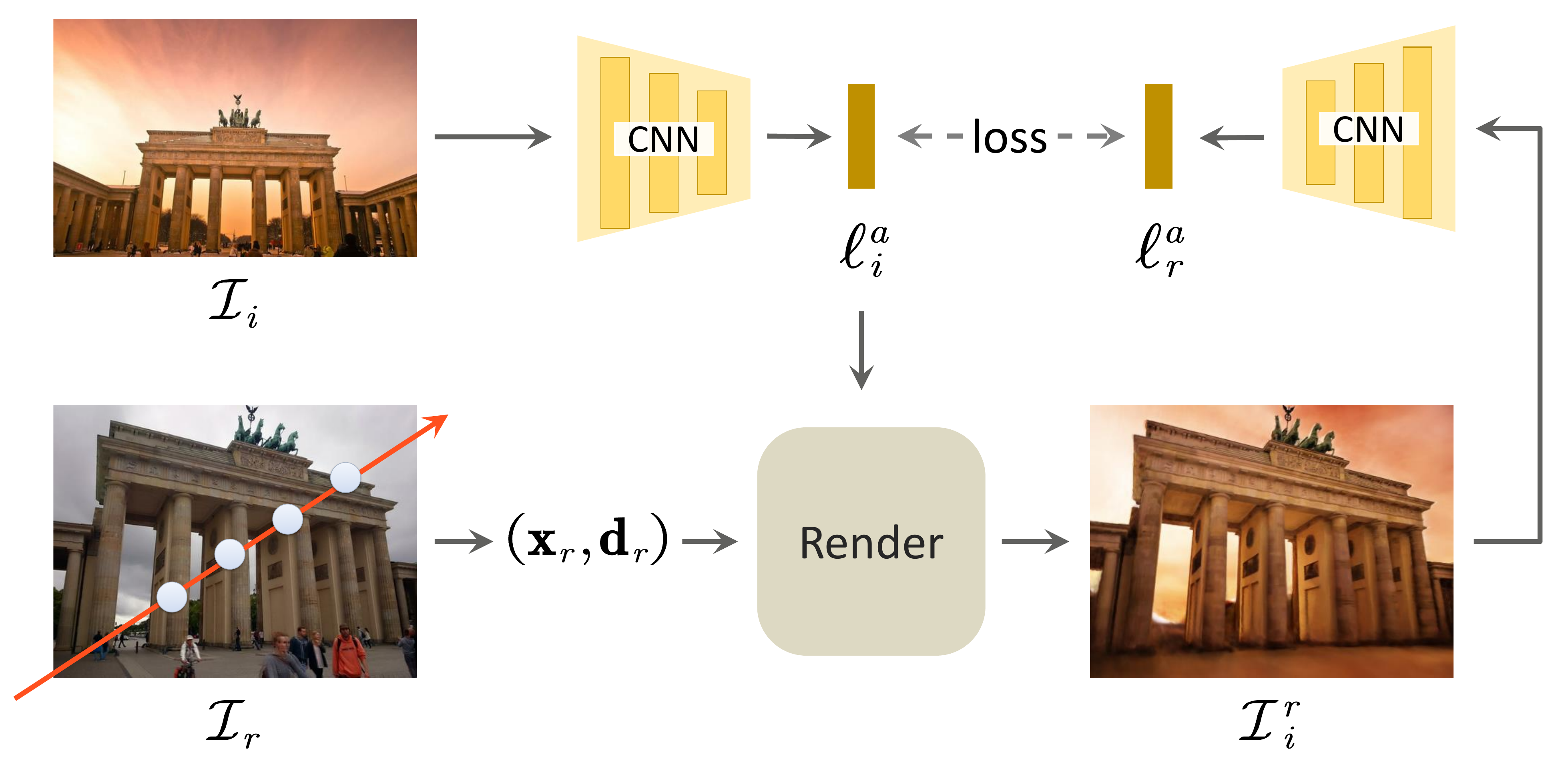}
  \vspace{-0.8cm}
   \caption{Illustration of view-consistent loss. Given an example image $\mathcal{I}_{i}$, we use a CNN to encode it into an appearance latent vector $\ell_{i}^{a}$. We sample camera rays in another view to render the hallucinated image $\mathcal{I}_{r}^{i}$ together with $\ell_{i}^{a}$. We encourage that the reconstructed appearance vector $\ell_{r}^{a}$ encoded from hallucinated image should be the same as $\ell_{i}^{a}$, since it is a global representation across different views.}
   \vspace{-0.6cm}
\label{fig:vcloss}
\end{figure}

However, the problem that disentangles the appearance from viewing direction with unpaired images is inherently ill-posed and requires additional constraints. Inspired by recent works \cite{zhu2017multimodal, huang2018multimodal, lee2018diverse} that exploit latent regression loss to encourage invertible mapping between image space and latent space, we propose a view-consistent loss $\mathcal{L}_{\mathrm{v}}$ to achieve the disentanglement of appearance and view by taking an appearance vector $\ell_{i}^{(a)}$ from the the appearance encoder $\operatorname{E}_{\phi}$ and attempt to reconstruct it in different views, which is formulated as:
\begin{equation}\small\label{eq9}
\mathcal{L}_{\mathrm{v}}=\left\|\operatorname{E}_{\phi}(\mathcal{I}^r_{i}) -\ell_{i}^{a}\right\|_{1},
\end{equation}
where $\mathcal{I}_{i}^{r}$ is the rendered image whose view is randomly generated and appearance is conditioned on the image $\mathcal{I}_{i}$ as shown in Fig.~\ref{fig:vcloss}. Here we assume that the reconstructed appearance vector $\operatorname{E}_{\phi}(\mathcal{I}_{i}^{r})$ should be the same as the original appearance vector $\ell_{i}^{a}$, since the appearance vector is a global representation across different views. Owing to the view-consistent loss, we can perform view-consistent appearance rendering, given the same appearance vector as input. In addition, we prevent encoding the image geometry content into the appearance vector with the help of view-consistent loss, which encodes the render images from different views (also content) to the same vector when conditioning the volume on the same vector.

To improve efficiency, we sample a grid of rays and combine them as the image $\mathcal{I}_{i}^{r}$ instead of rendering a whole image during training.~\cite{Schwarz2020NEURIPS}. This is based on the assumption that the global appearance vector of an image will remain unchanged after sampling using a random grid.

\subsection{Occlusion Handling}
\label{sec:occlusion}
Instead of using a 3D transient field to reconstruct the transient phenomena which is only observed in an individual image as in \cite{martin2021nerf}, we eliminate the transient phenomena using an image-dependent 2D visibility map. This simplification makes our method has a more accurate segmentation between the static scene and transient objects. To model the map, we employ an implicit continuous function $\operatorname{F}_{\psi}$ which maps a 2D pixel location $\mathbf{p}=(u, v)$ and an image-dependent transient embedding $\ell_{i}^{\tau}$ to a visible possibility $\mathcal{M}$:
\begin{equation}\small\label{eq12}
\mathcal{M}_{ij}=\operatorname{F}_{\psi}\left(\mathbf{p}_{ij}, \ell_{i}^{\tau}\right).
\end{equation}
We train the visibility map, which indicates the visibility of rays originated from the static scene, to disentangle static and transient phenomena of the images in an unsupervised manner with an occlusion loss $\mathcal{L}_{\mathrm{o}}$:
\begin{equation}\small\label{eq13}
\mathcal{L}_{\mathrm{o}}=\mathcal{M}_{ij}\left\|\mathbf{C}\left(\mathbf{r}_{ij}\right)-\hat{\mathbf{C}}\left(\mathbf{r}_{ij}\right)\right\|_{2}^{2}  + {\lambda_o}(1-\mathcal{M}_{ij})^2.
\end{equation}
The first term is the reconstruction error considering pixel visibility between the rendered and ground truth colors. Larger values of visible possibility $\mathcal{M}$ enhance the importance assigned to a pixel, under the assumption that it belongs to the static phenomena. The first term is balanced by the second, which corresponds to a regularizer with a multiplier $\lambda_o$ on invisible probability, and this discourages the model from turning a blind eye to static phenomena.

\begin{figure*}[htbp]
\centering
  \includegraphics[width=1\linewidth]{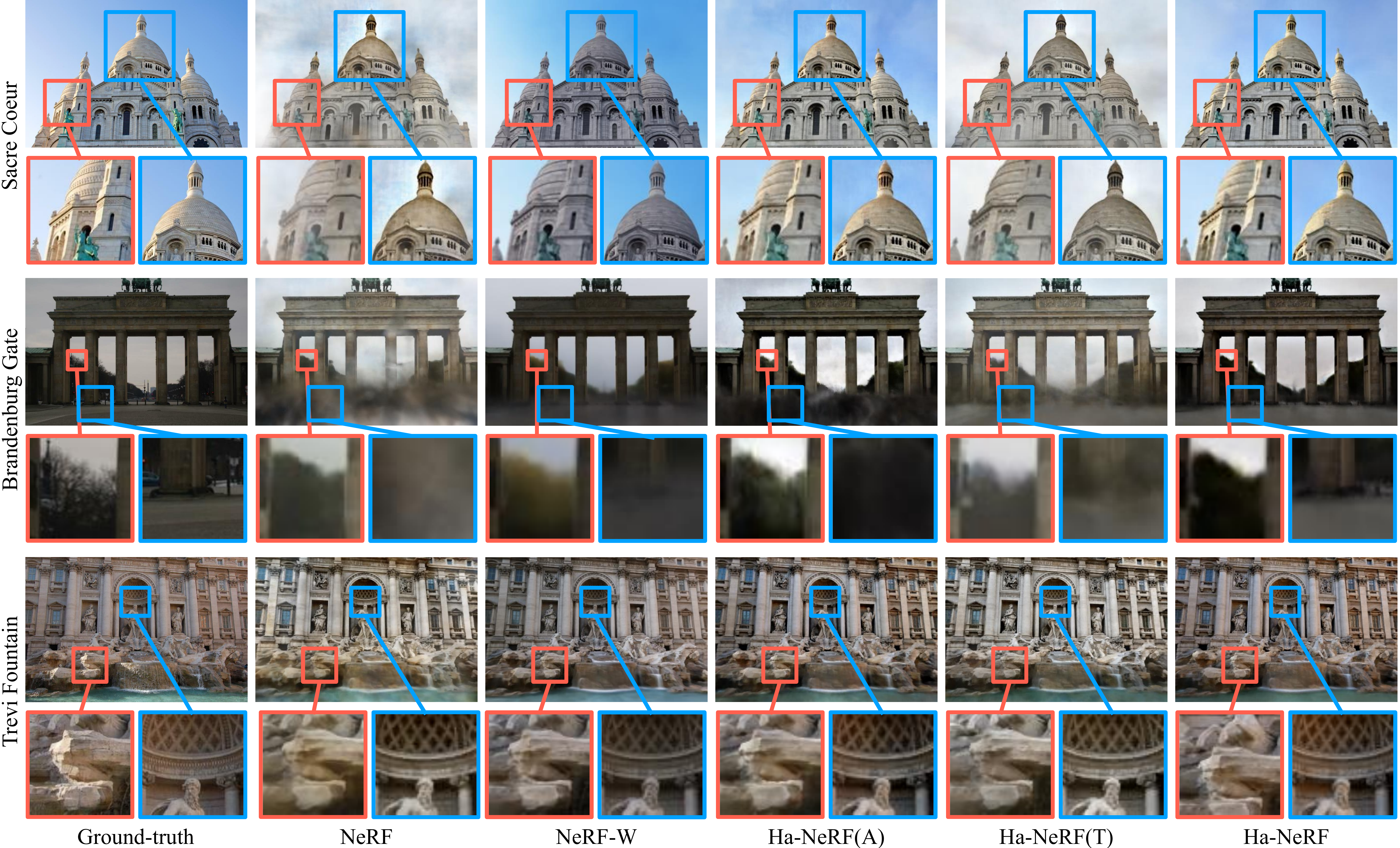}
  \vspace{-0.6cm}
   \caption{Qualitative results of experiments on constructed dataset. Ha-NeRF is able to encode the appearances and transfer them to novel views photo realistically (\eg, blue sky and sunshine in ``Sacre Coeur'', plants in ``Brandenburg Gate'', light reflection in ``Trevi Fountain''). Besides, Ha-NeRF removes transient occlusions to render a consistent 3D scene geometry (\eg, square and pillars in ``Brandenburg Gate'').}
   \vspace{-0.2cm}
\label{fig:exp_phototourism}
\end{figure*}

\begin{table*}[t]
\centering
\small
\begin{threeparttable}
\begin{tabularx}{1\textwidth}{@{}lCCCCCCCCC}
    \hline
    & \multicolumn{3}{c}{Brandenburg Gate} & \multicolumn{3}{c}{Sacre Coeur} & \multicolumn{3}{c}{Trevi Fountain} \\
    \cmidrule(lr){2-4} \cmidrule(lr){5-7} \cmidrule(lr){8-10}
    \multirow{-2}{*}{} & PSNR $\uparrow$ & SSIM $\uparrow$ & LPIPS $\downarrow$
                       & PSNR $\uparrow$ & SSIM $\uparrow$ & LPIPS $\downarrow$
                       & PSNR $\uparrow$ & SSIM $\uparrow$ & LPIPS $\downarrow$ \\ 
    \hline
    
    NeRF\cite{mildenhall2020nerf} & 18.90 & .8159 & .2316 & 15.60 & .7155 & .2916 & 16.14 & .6007 & .3662 \\

    NeRF-W\cite{martin2021nerf}* &  \cellcolor{best}{24.17} &  \cellcolor{best}{.8905} &  \cellcolor{second}{.1670} & 19.20 &  \cellcolor{best}{.8076} & .1915 &  18.97 &  \cellcolor{best}{.6984} & .2652 \\
    
    \hline
    
    Ha-NeRF(A) & 22.93 & .8517 & .1727 &  \cellcolor{second}{19.57} & .7864 &  \cellcolor{second}{.1839} & \cellcolor{second}{19.89} & .6798 &  \cellcolor{second}{.2377} \\
    
    Ha-NeRF(T) & 19.84 & .8368 & .1835 & 16.66 & .7657 & .2267 & 15.92 & .6186  & .2830  \\
    
    Ha-NeRF &  \cellcolor{second}{24.04} &  \cellcolor{second}{.8773} &  \cellcolor{best}{.1391} &  \cellcolor{best}{20.02} &  \cellcolor{second}{.8012} &  \cellcolor{best}{.1710} &  \cellcolor{best}{20.18} &  \cellcolor{second}{.6908} &  \cellcolor{best}{.2225} \\
    
    \hline
\end{tabularx}
\begin{tablenotes}
        \footnotesize
        \item[*] NeRF-W optimizes appearance vectors on the left half of each test image while Ha-NeRF does not.
\end{tablenotes}
\end{threeparttable}
\vspace{-0.3cm}
\caption{Quantitative results of experiments on our constructed dataset. Ha-NeRF achieves competitive PSNR and SSIM while outperforming the others on LPIPS across all datasets, even with the unfair experiment settings when compared with NeRF-W.}
\label{tab:real_scene}
\vspace{-0.5cm}
\end{table*}

\subsection{Optimization}
To achieve Ha-NeRF, we combine the aforementioned constraints and jointly train the parameters $\left(\theta, {\phi}, {\psi}\right)$ and the per-image transient embedding $\left\{\ell_{i}^{\tau}\right\}_{i=1}^{N}$ to optimize the full objective:
\begin{equation}\small\label{eq14}
\mathcal{L}=\lambda \sum_{i}\mathcal{L}_{\mathrm{v}} + \sum_{ij}\mathcal{L}_{\mathrm{o}}.
\end{equation}

\section{Experiments}
\subsection{Implementation Details}

Our implementation of NeRF and NeRF-W follows\cite{queianchen_nerf}. The static neural radiance field $\operatorname{F}_{\theta}$ consists of 8  fully-connected layers with 256 channels followed by ReLU activations to generate $\sigma$ and one additional 128 channels fully-connected layer with sigmoid activation to output the appearance-dependent RGB color $\mathbf{c}$. The appearance encoder $\operatorname{E}_{\phi}$ consists of 5 convolution layers followed by an adaptive average pooling and a fully-connected layer to get the appearance vector $\ell_{i}^{(a)}$ with 48 dimensions. The image-dependent 2D visibility mask $\operatorname{F}_{\psi}$ is modeled by 5 fully-connected layers of 256 channels followed by sigmoid activation to generate the visible possibility $\mathcal{M}$ conditioned on transient embedding $\ell_{i}^{\tau}$ with 128 dimensions. We set $\lambda$ to $1\times10^{-3}$ and $\lambda_o$ to $6\times10^{-3}$.

To evaluate the performance of Ha-NeRF in the wild, We constructed three datasets called ``Brandenburg Gate'', ``Sacre Coeur'' and ``Trevi Fountain'' using the Phototourism dataset, which consists of internet photo collections of cultural landmarks. We downsample all the images by 2 times during training.

\begin{figure*}[htbp]
\centering
  \includegraphics[width=1\linewidth]{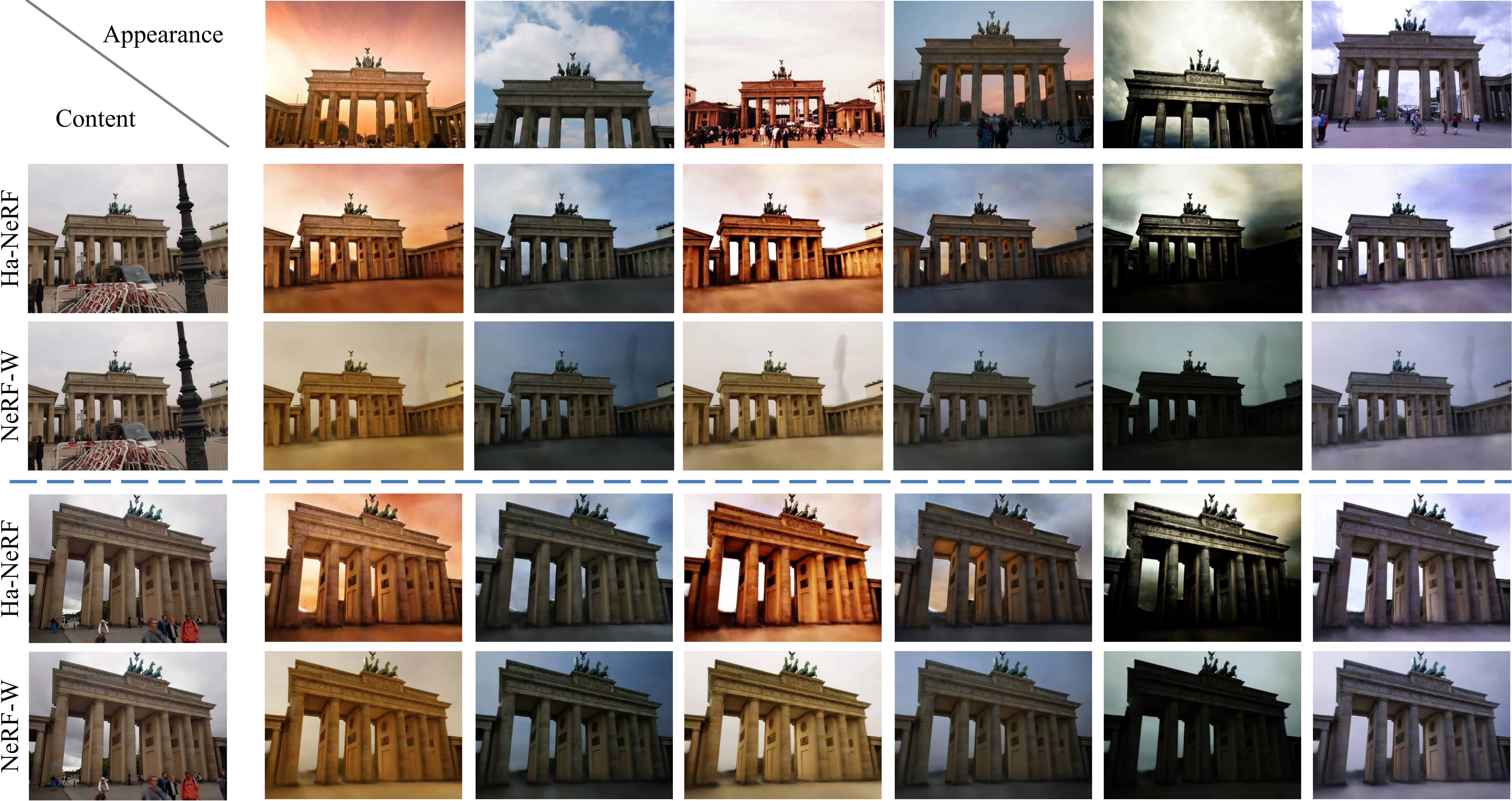}
  \vspace{-0.4cm}
   \caption{Hallucination in the ``Brandenburg Gate'' dataset with the global color shifts, such as weather, season and postprocessing filters. There are the images whose viewing direction is the same as the leftmost column content images, and the appearance is conditioned on the top line example appearance images.}
  \vspace{-0.4cm}
\label{fig:exp_transfer1}
\end{figure*}

\subsection{Evaluation}
\noindent\textbf{Baselines.}
We evaluate our proposed method against NeRF, NeRF-W, and two ablations of Ha-NeRF: Ha-NeRF(A) and Ha-NeRF(T). Ha-NeRF(A) (appearance) builds upon our full model by eliminating the visibility network $\operatorname{F}_{\phi}$, while Ha-NeRF(T) (transient) removes the appearance encoder $\operatorname{E}_{\psi}$ from the full model. Ha-NeRF is the complete model of our method.

\begin{figure*}[htbp]
\centering
  \includegraphics[width=1\linewidth]{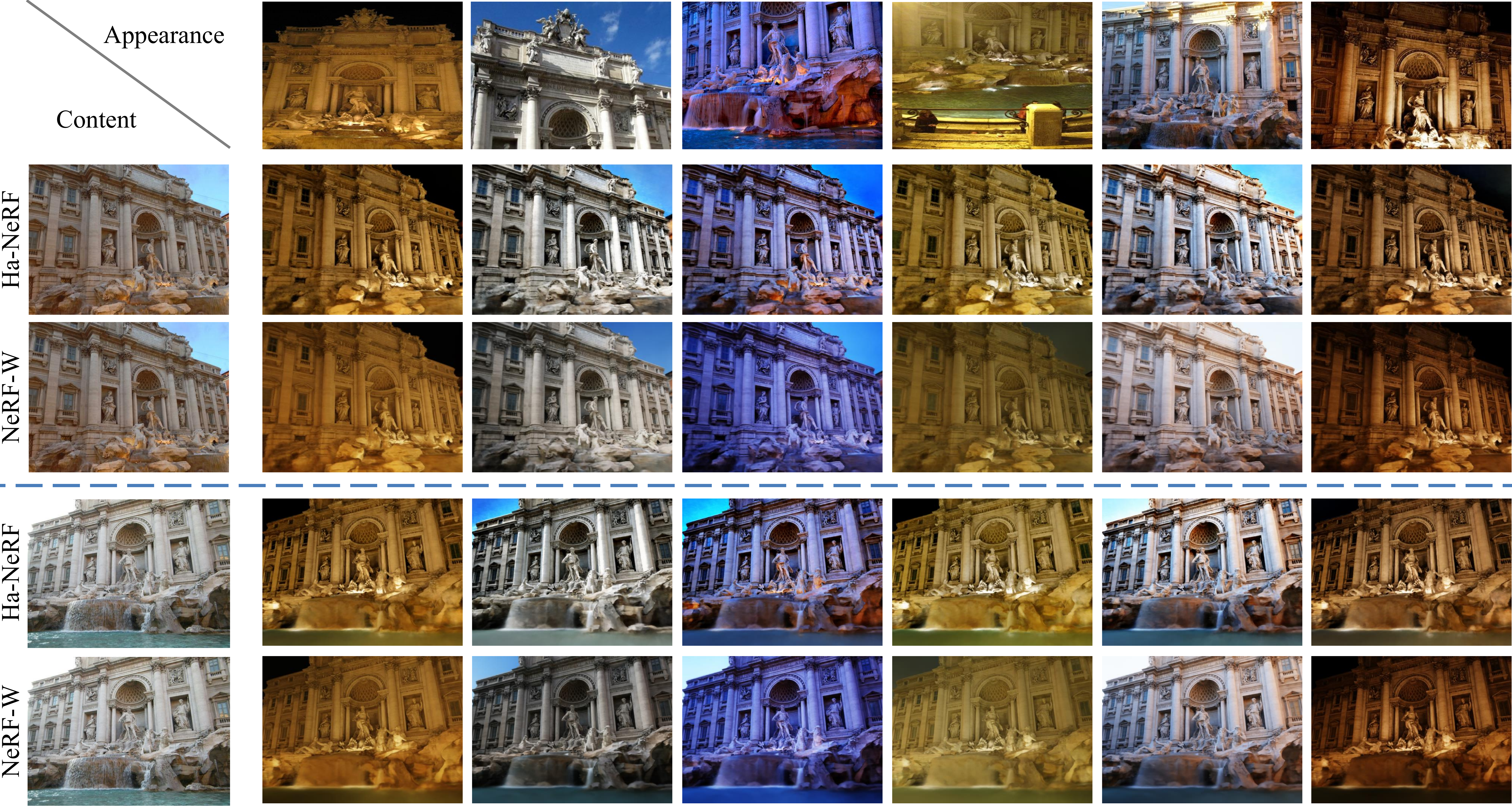}
  \vspace{-0.6cm}
   \caption{Hallucination in the ``Trevi Fountain'' dataset with high-frequency information of appearance, such as sunshine and colored light reflection. There are the images whose viewing direction is the same as the leftmost column content images, and the appearance is conditioned on the top line example appearance images.}
   \vspace{-0.4cm}
\label{fig:exp_transfer2}
\end{figure*}

\begin{figure*}[htbp]
\centering
  \includegraphics[width=1\linewidth]{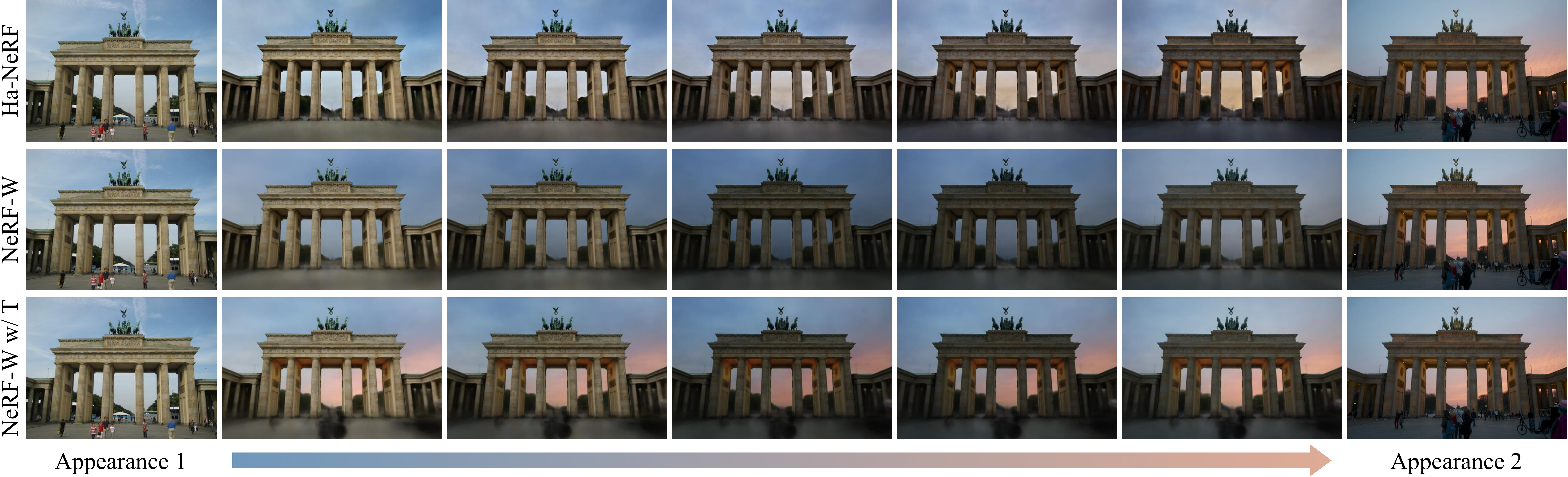}
  \vspace{-0.6cm}
  \caption{Images rendered from a fixed camera position with interpolated appearance between appearance $1$ and appearance $2$.}
  \vspace{-0.6cm}
\label{fig:interpolate}

\end{figure*}

\noindent\textbf{Comparisons.}
We evaluate our method and baselines on the task of novel view synthesis. All methods use the same set of input views to train the parameters and embedding for each scene except NeRF-W, which uses the left half of each test image to optimize the appearance embedding for the test set since they can not hallucinate new appearance without optimizing during training. We present rendered images for visual inspection and report quantitative results based on PSNR, SSIM, LPIPS.

Fig.~\ref{fig:exp_phototourism} shows qualitative results for all models and baselines on a subset of scenes. NeRF suffers from ghosting artifacts and global color shifts. NeRF-W produces more accurate 3D reconstructions and is able to model varying photometric effects. However, it still suffers from blur artifacts like the fog effect around the peristele of ``Brandenburg Gate''. This fog effect is the consequence of NeRF-W's attempt to estimate a 3D transient field to reconstruct the transient phenomena, while the transient objects are only observed in a single image. At the same time, renderings from NeRF-W also tend to exhibit different appearances compared to the ground truth, such as the sunshine and the blue sky in``Sacre Coeur'' and the light reflection in ``Trevi Fountain''.

Ha-NeRF(A) has a more consistent appearance, such as the blue sky at the top of ``Sacre Coeur''. However, it is unable to reconstruct high-frequency details due to the occlusion. In contrast, Ha-NeRF(T) is able to reconstruct structures with occlusion such as the square of ``Brandenburg Gate'', but is unable to model varying photometric effects. Ha-NeRF has the benefits of both ablations and thereby produces better appearance and anti-occlusion renderings.

    Quantitative results are summarized in Table \ref{tab:real_scene}. Optimizing NeRF on photo collections in the wild leads to particularly poor results that cannot compete with NeRF-W. In contrast, Ha-NeRF achieves competitive PSNR and SSIM compared to NeRF-W while outperforming the others on LPIPS across all datasets. Actually, this comparison is unfair to us. To transfer the appearance from test images, NeRF-W needs to optimize the appearance vectors on a subset of the test images during training. While Ha-NeRF does not use any test images during training. When testing, Ha-NeRF can directly encode the image appearance by a learned encoder. Despite this, our method still can produce competitive results compared with NeRF-W. 
    Moreover, NeRF-W exhibits view inconsistency. As the camera moves, renderings conditioned on the same appearance embedding appear to have an inconsistent appearance, which can not be reflected by current metrics.
    And we put the results into the supplemental material for the consistency comparison of NeRF-W with Ha-NeRF.

\noindent\textbf{Appearance Hallucination.}
By conditioning the color on the latent vector $\ell_{i}^{(a)}$, we can modify the lighting and appearance of a rendering without altering the underlying 3D geometry. In the meantime, encoding appearance with an encoder $\operatorname{E}_{\phi}$ allows our framework to perform example-guided appearance transfer. 

In Fig.~\ref{fig:exp_transfer1}, we see rendered images produced by Ha-NeRF using different appearance vectors extracted from example images. We also show the results of NeRF-W where appearance vectors are optimized during training. Notice that Ha-NeRF hallucinates realistic images while NeRF-W suffers from global color shifts compared with the example images. Moreover, Fig. \ref{fig:exp_transfer2} shows that Ha-NeRF can capture the high-frequency information of appearance and hallucinate the sunshine and colored light reflection of the scene.

Ha-NeRF can also interpolate the appearance vectors to get other hallucinations. In Fig.~\ref{fig:interpolate}, we present five images rendered from a fixed camera position, where we interpolate the appearance vectors encoded from the leftmost and rightmost images. Note that the appearance of the rendered images is smoothly transitioned between the two endpoints by Ha-NeRF. However, the interpolated results of NeRF-W completely ignore the sunset glow. Furthermore, we add the transient field of NeRF-W during its rendering (NeRF-W w/T), which shows the sunset glow. It reveals that NeRF-W could not disentangle the variable appearance (sunset glow) from transient phenomena (people) well.

\begin{figure*}[t]
\centering
  \includegraphics[width=1\linewidth]{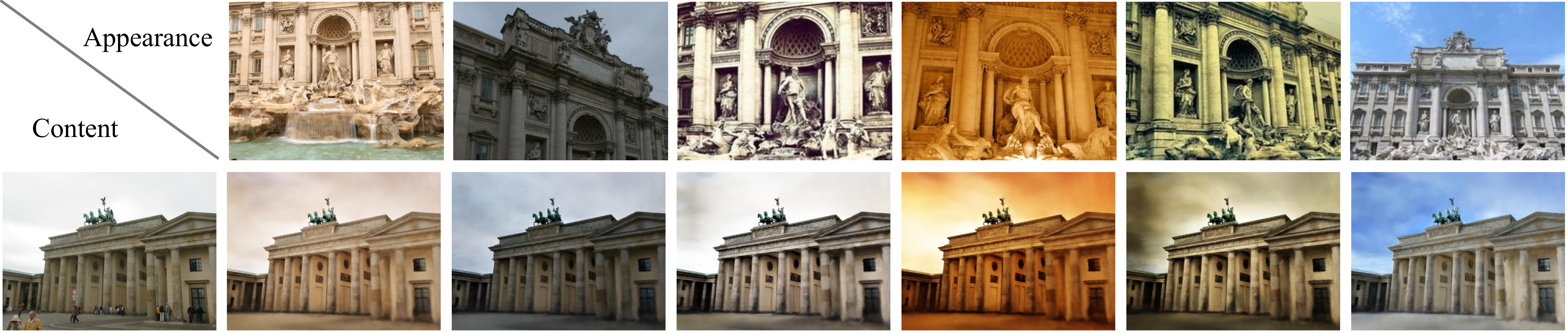}
    \vspace{-0.5cm}
   \caption{Cross dataset hallucination in the ``Brandenburg Gate'' condition on the example images of ``Trevi Fountain''.}
\label{fig:transfer_cross_2}
\vspace{-0.4cm}
\end{figure*}

\begin{figure}[t]
\centering
  \includegraphics[width=1\linewidth]{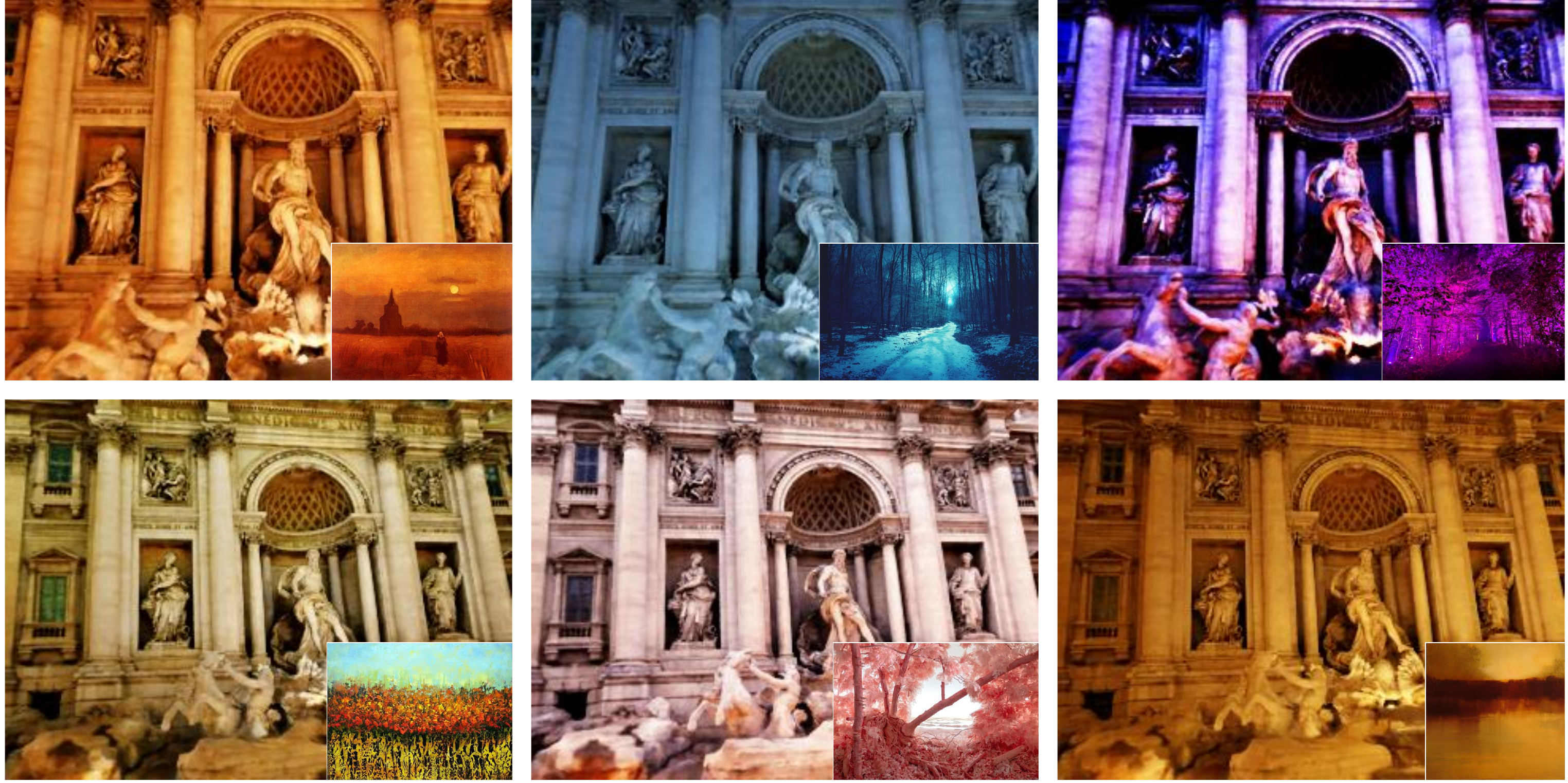}
    \vspace{-0.6cm}
   \caption{Hallucinating appearance from radically different scenes in different views. (\eg, forest artwork to building)}
\vspace{-0.6cm}
\label{fig:cross}
\end{figure}

\noindent\textbf{Cross-Appearance Hallucination.} We can perform appearance transfer by a user-provided example image from a different dataset. As shown in Fig.~\ref{fig:transfer_cross_2}, we hallucinate new appearance for ``Brandenburg Gate'' condition on the example image of ``Trevi Fountain''.
We can even transfer appearance from a radically different scene, as shown in \cref{fig:cross}, where there is a large domain gap between appearance images and scenes. We note that NeRF-W inherently can not hallucinate an appearance from other datasets because NeRF-W needs to optimize the appearance vectors on the example images, which must depict the same place.

\begin{figure}[t]
\centering
  \includegraphics[width=1\linewidth]{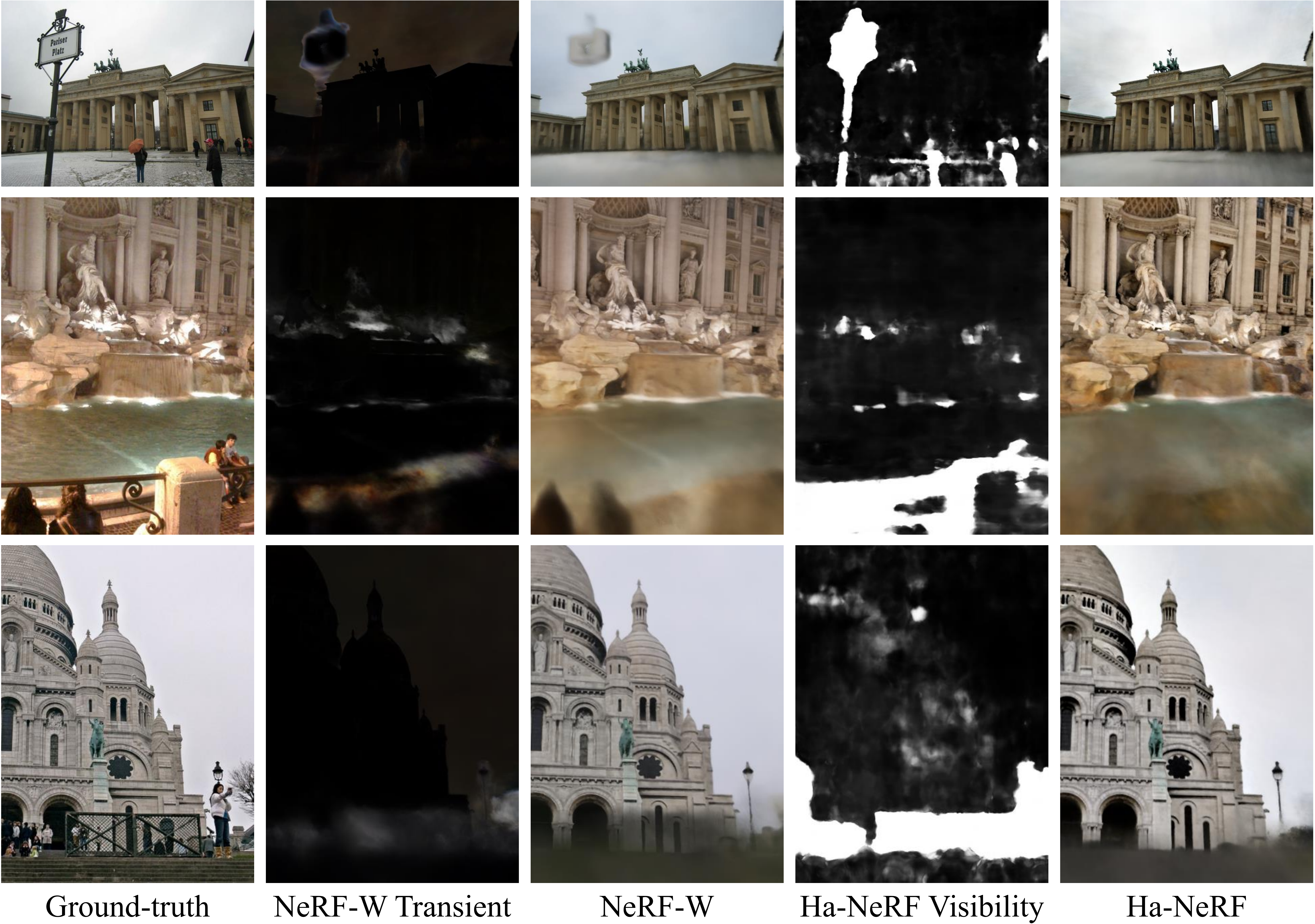}
   \caption{Anti-occlusion renderings of Ha-NeRF and NeRF-W. NeRF-W Transient is the renderings of the 3D transient field of NeRF-W, which tries to reconstruct the transient objects only observed in an individual image. We denote Ha-NeRF Visibility as our 2D visibility map that learned to disentangle static and transient phenomena of the images, indicating the visibility of rays originated from the static scene.}
\label{fig:mask}
\end{figure}

\noindent\textbf{Occlusion Handling.} 
We eliminate the transient phenomena using an image-dependent 2D visibility map, while NeRF-W uses a 3D transient field to reconstruct the transient objects. 
As illustrated in Fig.~\ref{fig:mask}, our occlusion handling method generates an accurate segmentation between static scene and transient objects, which allows us to render occlusion-free images. However, NeRF-W inaccurately decomposes the scene (\eg, board, people, and fence still leave on the renderings of NeRF-W) and further entangles the variable appearance and transient occlusion in the 3D transient field (\eg, results in the transient volume to remember the white cloud of ``Brandenburg Gate'').

\noindent\textbf{Limitations.} Without exception, the proposed Ha-NeRF suffers from the noisy camera extrinsic parameters, similar to most NeRF based approaches. Additionally, the quality of synthesized images degrades while the input images are either motion-blurred or defocused. Specific techniques have to be developed to handle these issues.

\section{Conclusion}

NeRF has grown in prominence and has been utilized in various applications, including the recovery of NeRF from tourism images. While NeRF-W works effectively with a train-data optimized appearance embedding, it is hard to hallucinate novel views consistently at an unlearnt appearance. To overcome this challenging problem, we present the Ha-NeRF, which can hallucinate the realistic radiance field under variable appearances and complex occlusions. Specifically, we propose an appearance hallucination module to handle time-varying appearances and transfer them to novel views. Furthermore, we employ an anti-occlusion module to learn an image-dependent 2D visibility mask capable of accurately separating static subjects. Experimental results using synthetic data and tourism photo collections demonstrate that our method can render free-occlusion views and hallucination of the appearance. Codes and models will be publicly available to the research community to facilitate reproducible research.

\newpage
{\small
\bibliographystyle{ieee_fullname}
\bibliography{egbib}
}

\end{document}